# NOVEL MACHINE LEARNING ALGORITHMS FOR CENTRALITY AND CLIQUES DETECTION IN YOUTUBE SOCIAL NETWORKS

Craigory Coppola and Heba Elgazzar

School of Engineering and Computer Science, Morehead State University
Morehead KY, USA

*ABSTRACT*

*The goal of this research project is to analyze the dynamics of social networks using machine learning techniques to locate maximal cliques and to find clusters for the purpose of identifying a target demographic. Unsupervised machine learning techniques are designed and implemented in this project to analyze a dataset from YouTube to discover communities in the social network and find central nodes. Different clustering algorithms are implemented and applied to the YouTube dataset. The well-known Bron-Kerbosch algorithm is used effectively in this research to find maximal cliques. The results obtained from this research could be used for advertising purposes and for building smart recommendation systems. All algorithms were implemented using Python programming language. The experimental results show that we were able to successfully find central nodes through clique-centrality and degree centrality. By utilizing clique detection algorithms, the research shown how machine learning algorithms can detect close knit groups within a larger network.*

*KEYWORDS*

*Social Networks, Maximal Cliques, Centrality, Machine Learning, Network Analytics, Clustering*

## 1. INTRODUCTION

The rise of social networks has created an influx of data that describes how individuals are connected. This data is easily accessible through web crawling, publicly available and large scale. Such large-scale datasets allow for algorithms from graph theory and machine learning to take hold and become extremely useful [1]. Analysis of network topology and subgroups can be used to find circles of people within a social network, or to recommend new friends given your current friends. One key use for social data can come in the form of commercial benefits.

As more and more data has become available, it has become possible to tailor advertisements to individuals' interests. Currently this takes the form of keywords being assigned to an individual. This can be seen as recently the social media network Instagram [2] has recently enabled their users to see what tags have been assigned to them. Also, advertising networks are looking to individuals to help advertise their products via sponsorship. This normally takes the form of paid product placement during a viewing, or posts which promote the product or company. Identifying who to sponsor is an interesting problem that companies face, as sponsors with little reach may end up costing the company more money than they bring in through new customers. In this paper, methods are proposed to help solve both issues. Particularly, clustering is used to take the place of keywords in identifying advertising demographics, and key figure analysis as a method to determine sponsorship figures.

This is made possible due to the large amount of data that can be captured from these "social sources." With users performing several actions per session, with those actions all being recorded data mining can easily be applied to gather large amounts of data from social sites. When talking about big data, it is important to discuss the "5 V's of big data" [3]. In particular, these are volume, velocity, variety, veracity, and value. Volume and velocity are straight forward, referring to the size of the data and the speed at which it is gathered and processed. Variety means that the data is often complex, with many features. Veracity refers to the fact that frequently data might have missing pieces, or holes, that must be dealt with in processing. Value simply refers to the value that is created by analyzing big data.

Given social data, it is easy to locate these 5 V's. The volume refers to the size of the data. In a social setting such as YouTube, Twitter, Facebook etc. this is caused by the sheer number of accounts that are held on each website. The velocity comes from the frequency of user interactions on the platform. A video might get a new view every 10 seconds, and a like every 15. Over the millions of videos, this results in a rapidly increasing dataset. A lot of social data sets might have some holes or missing pieces in it as user profiles might be incomplete. The value comes from the potential for analysis presented by the data. This is explained more in depth throughout the paper, as the results are discussed.

With big data, machine learning is even more useful [4]. New software and hardware tools such as open machine learning frameworks such as SciKit-learn [5], TensorFlow [6], and fast CPUs and GPUs allow easy entry to the field. This combination allows new companies specializing in the gathering of data and analyzing it with machine learning techniques to sprout and grow quickly.

While the context of our data set restricts our results to advertising suggestions, or influencer identification, these techniques could easily be applied to other datasets. By applying similar techniques to medical data, it would be possible to identify disease clusters. This would potentially allow easier identification of strains of the disease, and it would be possible to perform key-figure analysis on the clusters to find likely patient zeros. In this case, identifying the first patient can allow for research into cures to progress quicker as the disease progression can be monitored.

Given a data set of people with terroristic tendencies, similar techniques could identify leaders in the population, or provide a good starting point to look into certain individuals not previously known as a potential threat. While these applications may require tweaking of our proposal, these algorithms could easily be used as the base for the systems.

This work contributes a new and novel approach to utilize machine learning algorithms in graph analysis, specifically focusing on clustering and key figure analysis. With the use of Louvain modularity and spectral clustering, the work proposes that embedded clusters can be determined for the purpose of advertising. Key figure analysis utilizing multiple centrality measures is used to identify leaders of a community, and a proposal of using these measures within embedded communities is made.

The following sections will provide an overview of machine learning, specifically related to data clustering and centrality analysis, as well as a more in-depth discussion of the techniques used in these experiments. Previous work in the area will first be discussed, with a thorough discussion of what separates the prior work from this work. Ideas for progressing this research in the future will also be detailed, along with the explicit results from the current iteration of the research project.

## 2. RELATED WORK

Many other researchers have worked in this area, generally focusing on one part of the problem rather than both. Most papers have focused on developing new metrics for clustering in the graph-based network, but a few have worked on centrality. Very few papers have previously looked at both, and that lends to the novelty of the approach discussed in this paper.

Tang, Wang, and Liu [7] focus on connecting the multidimensional dataset for the purpose of community detection in this paper. They attempt to maximize the modularity of the network graph in an effort to locate the sub communities. This is not dissimilar to our approach with Louvain community detection but will be a harder computation. It should generate better results for sub-communities in the network datasets when multiple axis is clear and available but doesn't work towards identifying key figures or leaders within the subgroups at all.

Tang and Liu [8] also take a deeper dive into using modularity to analyze multidimensional networks. They utilize principle modularity maximization (PMM) to perform their community detection after combining the various dimensions into a workable dataset. This still focuses solely on the community detection and fails to regard important people and their influence in the community. However, this research makes use of the same YouTube dataset [9] that we use making it an important frame of reference for our work.

Agarwal and Yu [10] worked to find influencers among social media, but with a focus on bloggers. This is an important distinction, with bloggers interaction mainly being one-way. This differs from the YouTube dataset as we are looking at how the users interact with each other, and how they might share interests and interactions without directly going to look at the other user's page. As such, our methods are much better suited for "two-way" social datasets, compared to the method used by Agarwal which relies on the "one-way" nature of blogs.

Maharani, Andiwijaya, and Gozali [11] used similar centrality measures in their analysis of Twitter social media data. Their research was tailored towards overall centrality, on the basis that the 10 top influencers would be significantly different. In regard to our work, we work towards not only identifying influencers but limiting our search to particular advertising circles. This is different in that we are able to target a specific cluster and find a key figure within them, although similar analytic techniques are used.

Grando, Noble and Lab [12] performed similar analysis of centrality measures in regard to social networks. Their paper worked to distinguish which techniques would work well for finding central nodes. Taking this work into account, degree centrality was a good choice for finding key figures in communication networks. This is what one would be seeking when picking an advertising sponsor. While clique centrality is not mentioned, it is similar to subgraph centrality and mentioned as a good candidate for measuring community involvement and engagement. This is also good for the central purpose of the research.

Felipe, Noble, and Lab [13] also discuss the difficulty associated with computing vertex centrality for a large network. Their proposal of a neural learning model is useful for large scale datasets, but even so computing the measures we choose to use was not too computationally expensive. Enumeration of the cliques for the graph was completed prior to the use in centrality, and degree centrality was elementary.

Given a social network, applications for machine learning are abundant. The use of big data allows for patterns to be identified, and outcomes predicted [14]. Applications such as trust prediction [15] can be used to analyze political relationships, or elections. The insights that can be found can help identify relationships that might be dangerous in nature, with cells of activity being uncovered, but they can also help find new friends via features such as suggested friends. Machine Learning techniques can be used to suggest new videos, friends, and content creators. All of these uses lend to the credibility of the techniques, although privacy concerns create an atmosphere where one must be careful of how the data is used.

## 3. PROPOSED METHOD

The proposed methodology and the data flow of the used YouTube dataset is shown in Figure 1. For this research, unsupervised machine learning techniques were used to cluster nodes and identify key figures in our population. Supervised or reinforcement techniques could also be used for this purpose, with different algorithms. Unsupervised learning refers to the fact that the data is not pre-grouped, and no key figures were known before starting. Also, after the program began to run there was no external input. Had there been some data which was already determined to be in a cluster, or nodes that were known to be important supervised or reinforcement techniques could have been more effective. For implementing our algorithms, we used Python and the Sci-kit: Learn [5] library along with NetworkX [18], Matplotlib [19], and community. Using these libraries enabled us to focus on the results when combined with our data and ensures that the learning algorithms are correctly implemented. NetworkX [18] and Matplotlib [19] were used for representing the graphs in code and visually respectfully.

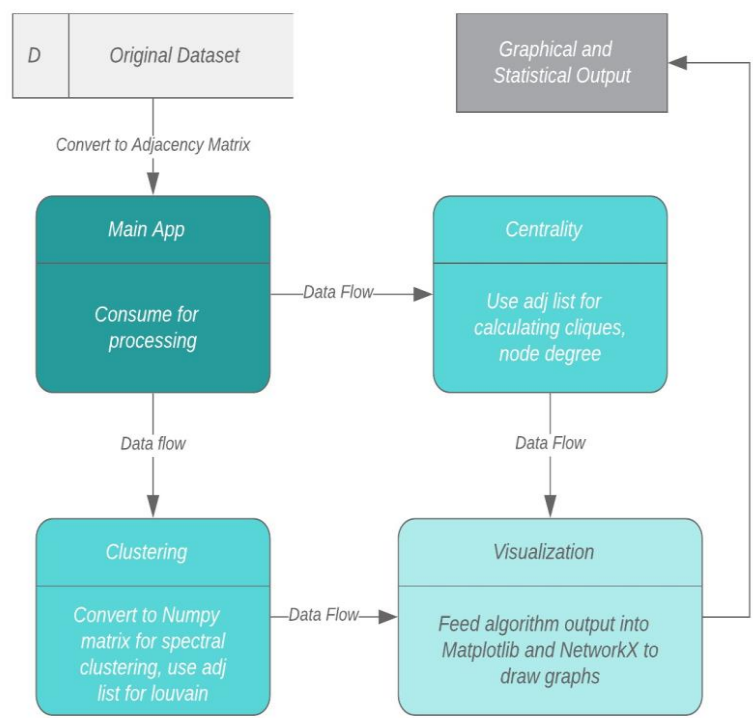

Figure 1. Proposed Methodology and the Data Flow for the YouTube Data

### 3.1.1. Clustering

Clustering algorithms partition data into groups known as clusters [16]. In the YouTube dataset, these clusters represent distinct sections of the community. After obtaining clusters, advertisers would be able to determine what the cluster has in common and broadcast ads to members of the cluster that are customized for that group. There are several methods for clustering which are frequently used, including K-means, spectral clustering, agglomerative clustering, DBSCAN, and affinity propagation. While these algorithms each have their own benefits, we focused on spectral clustering due to the ease of use with our dataset and cluster size.

### 3.1.2. Louvain Modularity

Louvain modularity is not strictly speaking a clustering algorithm, but it is a method to detect communities within a dataset. Given our purpose, to identify sub-communities for advertising, this algorithm also fits. Louvain modularity seeks to optimize the modularity of a graph. Modularity is the measure of the density of edges inside of the community versus edges the lead outside of the community [17]. This optimization can be represented as:

$$\frac{1}{2m} \sum_{ij} [A_{ij} - \frac{k_i k_j}{2m}] \delta(C_i C_j) \qquad (1)$$

In this equation, $C_x$ represents the community of node x. $A_{ij}$ represents the edge weight between nodes i and j. $K_x$ represents the sum of the weights of the edges attached to node x. 2m should be the sum of all weights in the graph. The open source Python Louvain [17] module was used to implement modularity for our dataset. This combined with NetworkX [18] and Matplotlib [19] created an easy implementation, that was only slowed by the visualization process given 15,088 nodes that had to be displayed discretely on a plane.

### 3.1.3. Spectral Clustering

One clustering technique is called spectral clustering. Spectral clustering operates on the similarity graph between nodes [20]. The graph's Laplacian is defined as L = D−W where D is the diagonal matrix with entries of each node's degree, and W is the weighed adjacency matrix. To generate a feature vector the algorithm calculates the first k eigenvectors for each node in the graph Laplacian. K-means or other clustering methods can then be used to separate our nodes into their prospective clusters.

Sci-kit Learn provides an implementation of this algorithm which we used to cluster our data [5]. In this implementation, either k-means or discretization is used to assign labels. Discretization was chosen to eliminate sensitivity to initialization states since that is controlled by the Sci-kit module.

### 3.1.4. Clustering Vs Community Detection

Both clustering and community detection algorithms work to find subgroups within a given dataset and are largely synonymous. They are both used contextually, depending on the algorithms used and what scientific community published the work. As such, both Louvain Community Detection and Spectral Clustering fall under the same category. The biggest distinction between the two (besides the manner in which the two algorithms operate) is that spectral clustering takes a parameter to determine the number of clusters. To get around this, machine learning algorithms ramp up the number of clusters and run multiple iterations while tracking metrics to determine the ideal number of clusters.

## 3.2. Centrality

Graph centrality when applied to human networks can help to locate influencers, or key figures within the population [21]. While this concept was first introduced in 1948 [22], it is still relevant in new networks. Companies sponsor key figures as a method of advertising their products, with content viewers being more likely to purchase something based on a review or mention from their favorite producers. This creates the concept of a Social Media Influencer. Online influencers have existed for a while now [23] with blogs and other review mediums being early targets for sponsorship. Social networks have allowed influencers to rise without being a purpose made review medium. On Instagram, Facebook, or YouTube a creator could find themselves approached by sponsors, when they originally where not planning to do sponsor spots or reviews. Centrality measures provide a method to identify these figures within a network, where previously companies either accepted applications or looked for channels that they knew about already.

With our dataset, centrality analysis is likely to find creators that have a heavy influence over the network. This provides an easy way for product creators to find new reviewers and potential sponsorship receivers for their products. While our results are based on the shared friends feature for centrality, this is likely to change based on your dataset and goal in the research. By looking at the shared friends features one will likely find the leaders of tightly knit groups, whereas shared subscriptions would be more likely to find the leaders of broad interest groups. We consider the following three measures of centrality:

### 3.2.1. Degree Centrality

Degree centrality assigns a numerical score to each node that represents the degree of the node [24]. This measure is relatively naive, with edge weighting not being taken in to account for our measures. While this means the strength of connections is not shown, it does show how many users an individual is contacting. Degree centrality scores are shown in Table IV-C.

### 3.2.2. Clique Centrality

Clique centrality assigns a numerical score to each node that represents the number of maximal cliques that include that node [25]. This makes it a measure of how many strongly connected groups that each node is a part of. For our discussion of influencers and key figures, this takes on a separate meaning. When looking for influencers, a company might care about how tightly knit the community is as the users might be more likely to listen to the producer.

A clique in a graph is a subset of nodes that forms a complete graph when considered on their own. This means every node in the subset should be connected to every other node in that set. The maximal cliques in a graph are all the cliques that are not part of a larger clique. As an example, in a complete graph there is only one maximal clique (containing all the nodes), since any subset of nodes is also fully connected to the nodes outside of that subset.

The well-known Bron-Kerbosch algorithm was used to locate the maximal cliques in our graph [26]. Bron-Kerbosch works via recursive backtracking. It takes 3 sets as input parameters:

- The nodes in the clique (initialized to ∅)
- The vertices remaining as possibilities (initialized to all nodes)
- The vertices not in the maximal clique (initialized to empty)

In general, Bron-Kerbosch works by considering all neighbors of an arbitrary element, e. For each neighbor, n, the algorithm considers the clique with e and n. If at any point there are no vertices that remain as a possibility, and there are also no vertices excluded the algorithm reports a maximal clique. Results for this measure are in table IV-C.

### 3.2.3. Average Rank Measure

The average score measure works to balance out the targeting of clique and degree centrality. Where in each of those methods a high score represents a central figure, lower scores are better here. This measure combines the results from the other methods, assigning a numerical score as: where Rx represents the rank of the node using x as scoring criteria. A generic formula for consider n ranks given other measures is:

$$\frac{\sum_{i=0}^{i<n}(R_i)}{n} \qquad (2)$$

where n is the number of ranks, and $R_i$ is the $i^{th}$ measurement rank.

By averaging out the ranks, nodes that scored highly in one consideration but poorly in another will be considered less heavily than nodes that did well in each method. This measure could be extended to include other criteria rankings if other scores had been calculated.

### 3.2.4. Degree Vs Clique Vs Average, And Other Methods

While this research focuses on Degree and Clique based centrality, other methods do exist and could prove suitable for key-figure analysis. Degree centrality was chosen due to its naivety, and obvious relevance when looking for the largest outreach an individual could have. In a less obvious way, clique centrality also achieves this. Clique centrality looks at how many nodes cliques a node is a part of, looking at the number of communities that a node is in rather than strictly how many nodes they can reach.

Other centrality measures that could be suitable for similar tasks include:

- Closeness Centrality: The closeness of two node is defined by the length of the shortest path between them. Closeness centrality sums these values for each node and marks it as the node's centrality score. Lower scores indicate more central nodes [27].
- Betweenness Centrality: The number of times that a node falls on the shortest path between two other nodes in the graph defines the original nodes betweenness score. A higher score indicates a more central node [28].
- Katz Centrality: Katz Centrality is similar to degree centrality, in that it looks at how many nodes are connected to the node that is being scored. However, Katz looks at the number of nodes that can be connected indirectly. This means that if the node can be connected via any number of other nodes, it still counts towards the original node's score. Higher values are more central nodes in this scoring method [29].

As other methods are evaluated, the Average Rank Measure becomes more attractive. This is due to its general formulation that allows it to account any number of rankings. When deciding your rankings, be sure to only include constructive rankings in the Average

## 4. EXPERIMENTAL RESULTS

### 4.1. Youtube Data

We refer to the Social Computing Data Repository from Arizona State University for our data [9]. This dataset is composed of 15,088 nodes crawled from the social network and video sharing website, YouTube. During the collection of data 848,003 user profiles were viewed. Each node represents a user, and for each user we are given five features. Our features are represented as edges between two nodes, and the weight of the edge represents the intensity of the interaction. Our five features are:

1) The contact network between the 15,088 users.
2) The number of shared friends between two users in the 848,003 viewed profiles.
3) The number of shared subscriptions between two users.
4) The number of shared subscribers between two users.
5) The number of shared favorite videos.

Data collection took place during December 2008. After downloading the data, several python scripts were implemented to interpret it. In particular, the format of the data was changed from [node, node, weight] to that of an adjacency matrix to be easier to analyze. This was the only preprocessing used before the data was fed into our algorithms. This does not include the preprocessing required to at run time, only that which was ran before any analysis began.

### 4.2. Methodology

Each metric output of our program consisted of a separate run, as the file did not do each in sequence. As such, the program was run a total of five times per feature, once for spectral clustering, Louvain modularity, and each of the three centrality methods. Each run consisted of pre-algorithm transformations, followed by the algorithm, followed by the creation of any visual output data. Along with the visual outputs, logs of the raw results were kept for analysis. More detail behind the preprocessing of our data, and the visual creation follows.

Matplotlib [19] was used when generating data visualizations. NetworkX' spring layout function was used to position the points on our graphs. This was necessary due to a lack of axis parameters in the data (i.e. there was no clear features to plot on an axis, as the features were relative to other nodes). Other layout methods were considered but resulted in too many nodes being hidden behind others. As such, when looking at data visuals remember that the location of the nodes is arbitrary, and not related to the data. Spring layout is meant to treat each edge in a network graph as a spring, with strength equal to the edge weight. Then it works to position each node such that the combined tension in those springs is minimal. This works well to generate clear visuals given a lower count of nodes but does cause some noise when presented with 15,000 nodes.

Python scripts were implemented to take in the raw data and initially convert it to an adjacency list format. That format is specified as (node, [(connected, weight)]). That is to say that for each node we store a list of its connections and weights. This format allows easy traversal and is used for the centrality measures outright. For clustering, Louvain modularity can work with this format. Spectral clustering however expected a NumPy Matrix, so the data was converted to one before the algorithm is ran.

With our dataset, it was easy to deal with any holes in the data that might have existed. This was due to the fact that we were looking at links in a graph. If a link was missing due to a data hole, it would simply be nonexistent in the constructed graph that was analyzed. If a large number of links were missing this would skew the results, but for our dataset the files were mostly complete

After results have been generated, the raw outcomes are immediately logged to a file. NetworkX [18] graphs are used to store the results in memory, before drawing the visualizations. A combination of Matplotlib [19] and NetworkX [18] is then used to generate graphs and save them to the disk. This allows one to look at the graphs or go back to the raw data and see the results.
This method of processing data and getting results is illustrated in Figure 1. Additionally, when calculating centrality data, links that had a strength of zero were dropped from our graph. This provided the means of interpreting the shared friends feature without it being a fully connected graph, which would not work for clique centrality or degree centrality (as neither consider edge weight).

### 4.3. Discussion Of Results

Figure 2 shows the output visualization for Louvain modularity. One of the issues faced in generating visualizations was the lack of location data for any of the nodes. The method used to determine the spatial placement of points was a mixture of NetworkX's spectral layout, and the spring layout methods. These methods take in a graph and try to generate coordinates that create good visuals. This caused long run times when generating visuals, with questionable output due to the lack of good axis. This does mean that the important part of our visuals lies in the coloration of the nodes not their positions. This is the case for each visual. Here, the color represents what community our Louvain modularity assigned each node to.

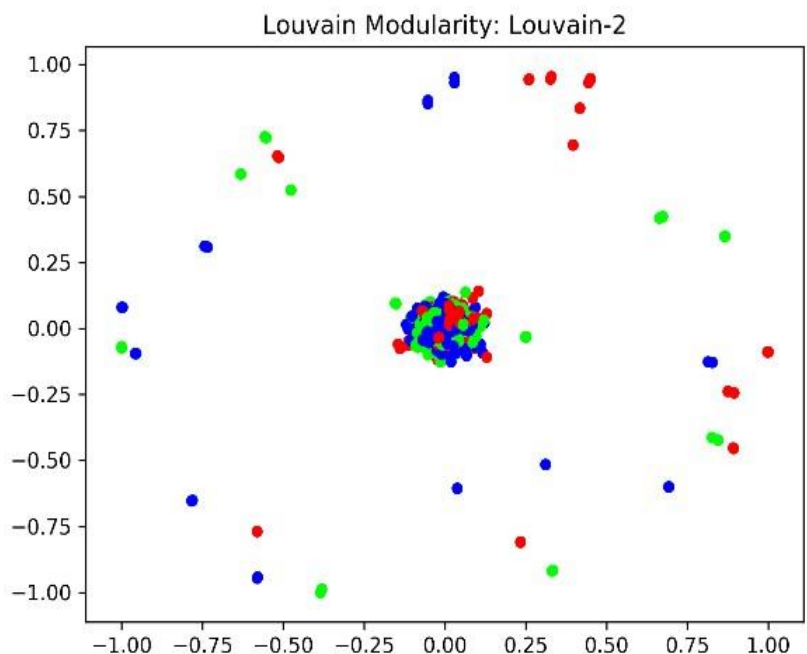

Figure 2. Proposed Methodology and the Data Flow for the YouTube Data

In Figure 3, the three identified clusters are displayed. This graph is a proof of concept that shows how these algorithms are able to perform given a social dataset. Utilizing additional metrics such as fit measures, one would be able to distill the actual number of clusters given a dataset. This is important for advertising but being able to set the suspected number of clusters is also valuable if one is looking for a certain number of community segments.

When considering which clusters to use, one can more or less just choose one. The results from the algorithms should be close to identical, despite the pictures looking so different in our cases. The reason the pictures look different here falls due to the lack of a good axis to place points. Both measures are applicable for the YouTube dataset, but perhaps Louvain modularity is a bit more suitable given the nature of the graph.

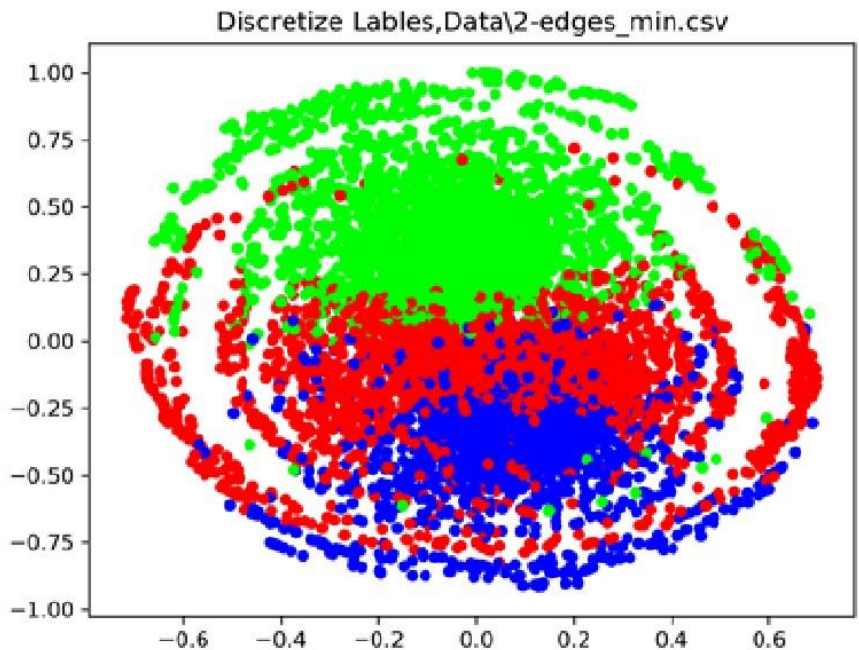

Figure 2. The three spectral clusters for the Shared Friends Feature

Table 1 shows the various centrality scores for the top-ranking nodes in our network. Given these results, it can be identified that Node 102 and 106 were particularly high ranking due to their top rankings in clique and average centrality. It would also be worth further investigating node 162 due to its top-ranking degree centrality score and high-ranking average score. It's important to investigate the metrics that are deemed appropriate for what one is currently attempting to identify. I.e., if one was looking for sheer outreach from a node, looking in the degree rankings would be more pertinent than the clique rankings. This would also be a case where it would be useful to consider what the feature one has chosen to analyze, with some features making more sense for a specific target use case.

When interpreting these results, it is important to realize that there are no ground truths for this data, as there is not in many real-world scenarios. If the ground truth was known, the algorithmic analysis would not be necessary. Due to this, verification of these results is left to statistical analysis. For centrality, this is easy to calculate by hand or proven scripts. On the other hand, the clustering and clique analysis is arduous and time consuming to prove but is shown to be accurate. When considering the three centrality scores, consider the use case for the data that you are looking at. If one wants someone with purely the most outreach, the degree centrality score will model that the most closely. If one wants members of a tightly knit group, the clique centrality will find leaders within those groups. As always, the explicit choice in features that one considers are based on the dataset that has been chosen. Given a more detailed dataset, one might choose to look more at information such as average post time, or connection strength rather than connection volume.

Table 1. The Centrality Results in The Format: Node (Score)

| Rank | Degree | Clique | Average Score |
|---|---|---|---|
| 1 | 162 (523) | 102 (6527) | 102 (4) |
| 2 | 4930 (502) | 106 (5331) | 106 (4) |
| 3 | 24 (407) | 110 (5057) | 110 (8) |
| 4 | 218 (404) | 6093 (4664) | 4926 (17) |
| 5 | 4971 (361) | 6091 (4290) | 162 (19) |
| 6 | 106 (338) | 5671 (3214) | 4930 (23) |
| 7 | 102 (311) | 5647 (2783) | 2581 (25) |
| 8 | 1189 (263) | 113 (2743) | 164 (31.5) |
| 9 | 3366 (253) | 6086 (2650) | 56 (33.5) |
| 10 | 170 (250) | 6085 (2415) | 4971 (35.5) |

For our results, we choose the shared friends feature due to those connections generally being stronger than a one-way feature such as shared subscribers vs shared subscriptions. This is evident if one considers the likelihood of nodes with high counts of shared friends vs shared subscribers vs shared subscriptions. Since subscribers and subscriptions are one-way cases, they are not as tightly knit as shared friends. An example of this behavior might be seen if you look at the social data from someone at the head of a large tech company. They might be followed by many of the same people as someone at the head of another company, but they likely have less shared friends when you look at people outside of their current company. In this case looking at shared friends might successfully cluster the company's employees as a cluster, while the shared subscribers feature would be more likely to cluster people with an interest in technology. This is also a good example of why context matters when performing the analysis. One of the limitations of this work is related to the dataset that was used. Additional datasets with new features may lead to better visuals and better analysis for other purposes.

## 5. CONCLUSIONS

Throughout this research a method for identifying key figures within our dataset was sought after, and an average rank measure proved to be the best, dependent on the situation. As with any problem, context matters. If one is seeking certain results, the measure to use will change. This same ideology is used with data clustering. Which method you use will be based on what results you are after. However, it holds that looking at the overlap between results provides a basis for identifying stable results.

This research provides a basis for analyzing social media datasets using Machine Learning with graph theoretic techniques. Means for analyzing data clusters and key figures within them have been found, and suitable metrics are also provided. This work could easily be applied to locating influencers and tight clusters within any given network. The clique analysis put forth gives a basis for finding leaders among smaller groups within a community, specifically leaders that partake in multiple groups. Similar techniques could also be used on this dataset, or other datasets for a variety of use cases depending on the analyzed feature.

Methods such as clustering and centrality analysis can be extremely useful in this area, especially due to the sheer volume of data that has become available. With many social network datasets, the features provided are plentiful with examples being geolocation, post contents, the posting time and the people that have left comments on them. By looking at the frequency of interactions one might be able to find even more insight.

There is much more work to be done in this area, with tailored algorithms being applicable on a per-dataset basis. This work could easily be translated to a medical scenario and testing such methods could prove useful for locating disease clusters and patient zero's in crisis situations. Other than varying the datasets, new measures such as Katz centrality [29] or weighted-degree centrality [30] could be tested to identify stronger classifiers when analyzing the dataset.